\documentclass[11pt]{article}
\usepackage[OT1]{fontenc}
\usepackage{coling2018}
\usepackage{times}
\usepackage{url}
\usepackage{latexsym}

\usepackage[small]{caption}
\usepackage{graphicx}
\usepackage{amsmath}

\title{Multi-Perspective Context Aggregation for \\ Semi-supervised Cloze-style Reading Comprehension}

\author{Liang Wang$^1$, Sujian Li$^2$, Wei Zhao$^1$, \\
  \textbf{Kewei Shen$^1$, Meng Sun$^1$, Ruoyu Jia$^1$, Jingming Liu$^1$} \\
  $^1$Yuanfudao Research, Beijing, China \\
  $^2$Key Laboratory of Computational Linguistics, Peking University, MOE, China \\
  {\tt \{wangliang01,zhaowei01,shenkw,sunmeng,jiary,liujm\}@fenbi.com} \\
  \tt lisujian@pku.edu.cn\\}

\begin{document}

\maketitle

\begin{abstract}
Cloze-style reading comprehension has been a popular task for measuring the progress
of natural language understanding in recent years.
In this paper,
we design a novel multi-perspective framework, which
can be seen as the joint training of heterogeneous experts
and aggregate context information from different perspectives.
Each perspective is modeled by a simple aggregation module.
The outputs of multiple aggregation modules are fed into a one-timestep pointer network to get the final answer.
At the same time, to tackle the problem of insufficient labeled data,
we propose an efficient sampling mechanism to automatically generate more training examples
by matching the distribution of candidates between labeled and unlabeled data.
We conduct our experiments on a recently released cloze-test dataset CLOTH~\cite{xie2017large},
which consists of nearly $100k$ questions designed by professional teachers.
Results show that our method achieves new state-of-the-art performance over previous strong baselines.
\end{abstract}

\section{Introduction}
\blfootnote{

    \hspace{-0.65cm}  
    This work is licensed under a Creative Commons
    Attribution 4.0 International License.
    License details:
    \url{http://creativecommons.org/licenses/by/4.0/}
}
Reading comprehension is a challenging task which requires the deep understanding of natural language.
Cloze test is a particular form of reading comprehension:
given a passage with blanks,
an examinee is required to fill in the missing word (or phrase) that best fits the context surrounding the blank.
Recently,
cloze-style reading comprehension has drawn growing interests from NLP research communities,
since such a task meets the practical requirements and is relatively easy to design.

The research of cloze-style reading comprehension is first advanced by two large-scale corpora:
the CNN/Daily Mail ~\cite{hermann2015teaching} and CBT ~\cite{hill2015goldilocks} datasets,
which are automatically constructed by randomly or periodically deleting a word from original passage.
Though the automatically generated datasets usually consist of a large quantity of labeled data
and make it possible to train large neural network models,
they are in nature far away from real-world language understanding problems
and have serious ambiguity issues~\cite{chen2016thorough}.
As a result,
the state-of-the-art system of cloze test almost reaches the performance ceiling
and loses its improvement direction due to the limitation of the corpus~\cite{chen2016thorough}.
In such situation,
~\newcite{xie2017large} argues
that it is a more reliable means to assess language proficiency
with carefully designed questions by professional teachers,
and releases a novel corpus CLOTH.
The CLOTH dataset brings the new challenge of exploring a comprehensive evaluation of language proficiency
and specifically divides the questions into several types including matching, reasoning and grammar etc.
Table ~\ref{table:question_type} shows several example questions from CLOTH.

\begin{table}[ht]
  \centering
  \scalebox{0.9}{\begin{tabular}{l|l}
      \hline
      \multicolumn{1}{c|}{question}  & \multicolumn{1}{c}{type} \\ \hline
      \begin{tabular}[c]{@{}l@{}}...... As a Senior student, I have to \underline{\hspace{0.4cm}} many exams. ......\\ \textit{A: finish\ \ \ \ B: win\ \ \ \ \textbf{C: take}\ \ \ \ \ D: join}\end{tabular}     & collocation       \\ \hline
      \begin{tabular}[c]{@{}l@{}}I am calling from the \underline{\hspace{0.4cm}} station ....... \\ `` There was an accident, and a man died .'' ......\\ \textit{A: post\ \ \ \ \ \ B: bus\  \ \ \ \ \textbf{C: police}\ \ \ \ D: railway}\end{tabular}   & matching  \\ \hline
      \begin{tabular}[c]{@{}l@{}}a student reported that I made an error ...... He was \underline{\hspace{0.4cm}} \\ and after thanking him for his honesty ...... he said angrily .\\ \textit{A: wise\ \ \ \ \ \ \textbf{B: right}\ \ \ \ C: rigid\ \ \ \ D: angry}\end{tabular} & reasoning \\ \hline
            \begin{tabular}[c]{@{}l@{}}...... They are used to \underline{\hspace{0.4cm}} messages by computers \\ and smart phones. ......\\
            \textit{\textbf{A: sending}\ \ \ B: send\ \ \ \ C: sent\ \ \ \ D: sends}\end{tabular}   & grammar   \\ \hline
    \end{tabular}}
    \caption{Example questions and their corresponding types from the CLOTH dataset.
    ``......'' represents some omitted irrelevant sentences.}
    \label{table:question_type}
\end{table}

From experiments by ~\newcite{xie2017large},
we can see that the \emph{Stanford attention reader}~\cite{chen2016thorough} of having the near state-of-the-art performance
(with an accuracy of about $0.74$) on CNN/Daily Mail
only gets an accuracy of $0.487$ on CLOTH
and there exists a huge performance gap between human and popular machine learning models.
The main reason is that attention models are mainly good at processing matching questions
(e.g., the first example in Table~\ref{table:question_type} has matching between \emph{``police''} and \emph{``accident''}, \emph{``man died''}),
which occupy a less percentage in CLOTH than in CNN/Daily Mail.
~\newcite{xie2017large} also present
the word-predicting potential of language models (LM) which can well tackle lexical collocation
(e.g., \emph{``take''} and \emph{``exam''} in the second example),
given a large volume of unlabeled data and high computation power.
Furthermore, ~\newcite{xie2017large} points out that the most difficult questions belong to the long-term-reasoning type (e.g., the third example question),
which constitutes approximately $22.4\%$ in CLOTH and needs more semantics to deal with.

To comprehensively consider the progress and questions in CLOTH,
we come up with the idea of modeling multiple perspectives to arrive at the correct answer,
given limited computation power.
Our multi-perspective network consists of several parallel modules,
where each module aggregates context information from a unique perspective.
We model long-distance matching with attentive readers,
global semantics with iterative dilated convolutions
and lexical collocation with both n-gram and neural language model(LM).
The outputs of aggregation modules are further integrated
and fed into a one-timestep pointer network~\cite{vinyals2015pointer} to get the final answer.

Next, one challenging problem is how to effectively train our multi-perspective network due to the insufficiency of labeled data.
To overcome this problem,
~\newcite{xie2017large} present a representativeness-based weighted loss function.
Their approach has two drawbacks:
first,
it requires to train another model for predicting a candidate's representativeness score;
second,
it is not a sample-efficient way
since each word including uninformative stop words becomes a training example.
In this paper,
we improve on ~\newcite{xie2017large}'s approach and develop a semi-supervised learning method by matching the distribution of candidates between labeled and unlabeled data.
The intuition is to make automatically constructed data as similar as possible to existing labeled data.
Stop words, named entities and out-of-vocabulary words should be downsampled
while meaningful content words should be kept for training.

Our method is simple, straightforward
and shows better performance with only a fraction of training examples.
Experiments show that our semi-supervised multi-perspective network is able to outperform state-of-the-art results on the CLOTH dataset
by $4.2\%$.

\section{Model}

Formally, the task of cloze-style reading comprehension requires
choosing the correct answer from $|c|$ candidates $\{c_i\}_{i=1}^{|c|}$
given a sequence of words $\{w_i\}_{i=1}^n$ as context.
Candidate $c_i$ could be a word or a phrase.
For the CLOTH dataset,
each question has $|c| = 4$ candidates.

\subsection{MPNet: Multi-Perspective Context Aggregation Network} \label{section:mpnet}

\begin{figure}[ht]
\begin{center}
 \includegraphics[width=0.55\linewidth]{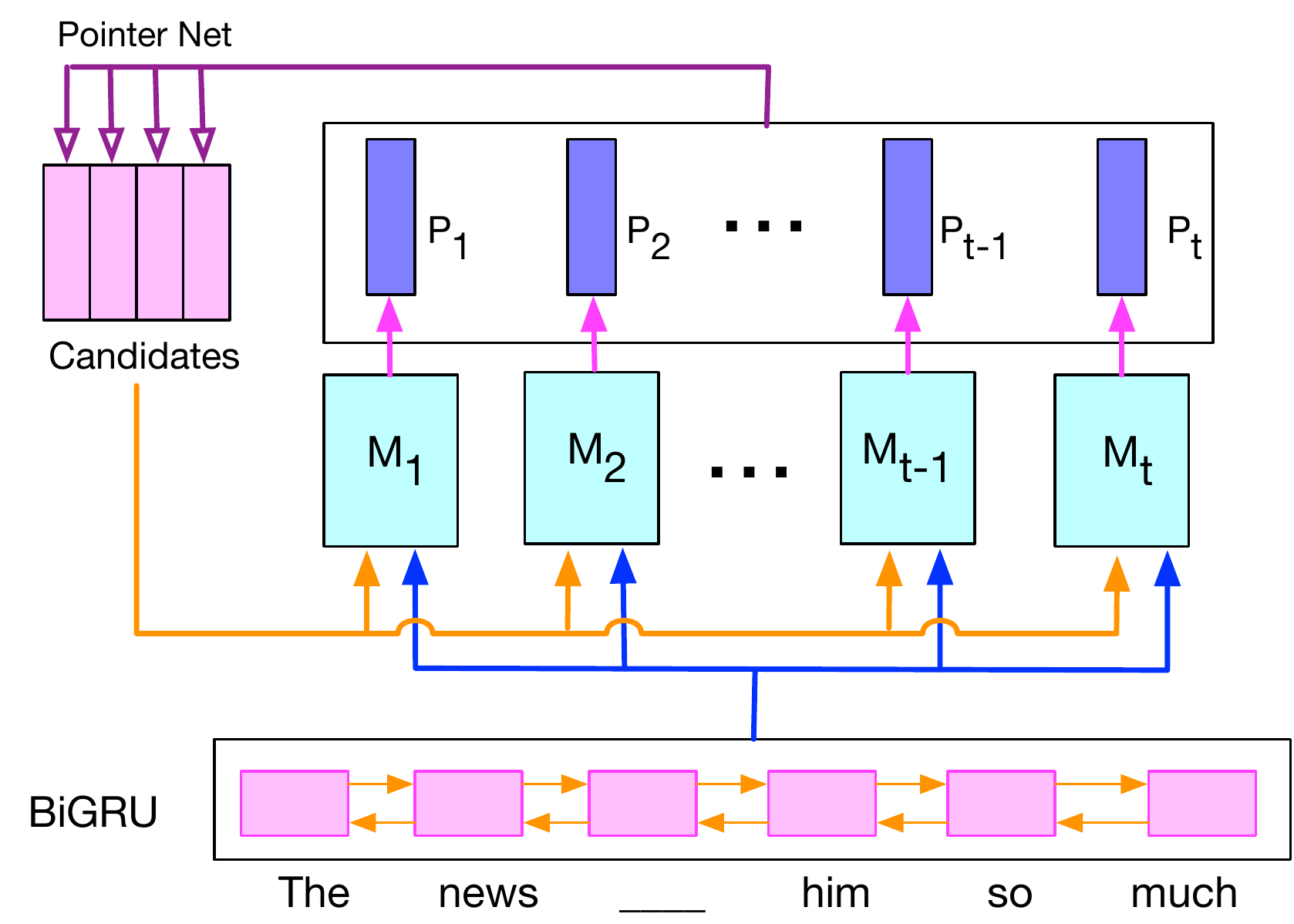}
 \caption{MPNet: \textbf{M}ulti-\textbf{P}erspective context aggregation network.
 We only show part of the context \emph{``The news \underline{\hspace{0.4cm}} him so much''}
 and ``\underline{\hspace{0.4cm}}'' is the blank to fill in.
}
 \label{fig:model}
\end{center}
\end{figure}

The overall architecture of our proposed MPNet is shown in Figure ~\ref{fig:model}.
It consists of an input layer,
a multi-perspective aggregation layer
and an output layer.

\noindent
\textbf{Input Layer\ \ }
Given the passage as a variable-length word sequence $\{w_i\}_{i=1}^n$,
we embed each word into $300$-dimensional word embeddings $\{\mathbf{e}_i\}_{i=1}^{n}$
using GloVe vectors.
Then, we apply bidirectional GRU(BiGRU) on $\{\mathbf{e}_i\}_{i=1}^{n}$
to get contextualized word representations $\{\mathbf{h}_i\}_{i=1}^n$ ~\cite{mccann2017learned} ~\cite{peters2018deep},
since GRU is computationally more efficient and shows slightly better performance than LSTM.
\begin{equation}
\begin{aligned}
    & \overrightarrow{\mathbf{h}}_i = \overrightarrow{GRU}(\overrightarrow{\mathbf{h}}_{i-1}, \mathbf{e}_i) \\
    & \overleftarrow{\mathbf{h}}_i = \overleftarrow{GRU}(\overleftarrow{\mathbf{h}}_{i+1}, \mathbf{e}_i) \\
    & \mathbf{h}_i = [\overrightarrow{\mathbf{h}}_i; \overleftarrow{\mathbf{h}}_i]
\end{aligned}
\end{equation}

We also use another GRU to encode candidates $\{c_i\}_{i=1}^{|c|}$ into fixed-length vectors $\{\mathbf{u}_i\}_{i=1}^{|c|}$,
as candidates may be multi-word phrases.
\newline

\noindent
\textbf{Multi-Perspective Aggregation Layer\ \ }
This layer consists of several independent aggregation modules $\{M_i\}_{i=1}^t$.
Computation can be easily parallelized since modules are independent.
Each module $M_i$ takes contextualized word representations $\{\mathbf{h}_i\}_{i=1}^n$
and candidates' encoding $\{\mathbf{u}_i\}_{i=1}^{|c|}$ as input
and outputs a vector $\mathbf{p}_i$,
which reflects the information from module $M_i$'s perspective.
We also assume aggregation modules can have access to $\{w_i\}_{i=1}^n$ and $\{c_i\}_{i=1}^{|c|}$.
For cloze-style reading comprehension,
each module should be able to distill some knowledge which can judge whether a candidate fits a given context from a perspective.

The aggregation modules that we use are listed below:
\begin{itemize}
\item \textbf{Selective Copying\ }
Assuming the index of the blank is $j$,
this module simply selects the hidden representation of the blank $\mathbf{h}_j$,
directly copies it to the output $\mathbf{p}_{sc}$ and ignores everything else.
Note that $\mathbf{h}_j$ is the output of BiGRU
and already incorporates context information from both forward and backward directions.
This resembles a bidirectional language model without softmax output layer.
Words near the blank are paid more attention which is consistent with our intuition of filling in the blank.

\item \textbf{Attentive Reader\ } \label{section:ar}
A large portion of questions involve matching candidates with related words
which may be far away from each other such as the second example in Table ~\ref{table:question_type}.
\emph{Attentive reader} proposed by ~\newcite{chen2016thorough} directly attends to the entire context
and therefore avoids the difficulty of modeling long-range dependence.
Original bilinear attention function ~\cite{chen2016thorough} is slightly modified
by introducing $\mathbf{b}_{ar}$ to model attention bias towards the $i$th word.
$\mathbf{u}$ is the vector representation of a candidate,
we omit its subscript for simplicity.
\begin{equation}
\begin{aligned}
    & \alpha_i = \mathrm{softmax}_i(\mathbf{u}^T\mathbf{W}_{ar}\mathbf{h}_i + \mathbf{b}_{ar}^T\mathbf{h}_i), i \in [1,n] \\
    & \mathbf{p}_{ar} = \sum_{i=1}^{n}\alpha_i \mathbf{h}_i
\end{aligned}
\end{equation}

\item \textbf{Iterative Dilated Convolution\ }
Convolutional neural networks have been a successful method
for modeling both natural language~\cite{kim2014convolutional} and images.
Multiple layers of CNNs can extract features in a hierarchical way,
which shares similarity with the compositional property of natural language.
Dilated convolution is a variant of traditional convolution
and is more efficient for multi-scale context aggregation~\cite{yu2015multi,strubell2017fast}.
In this work,
we use two blocks where each block consists of two dilated convolutions with dilation rate set to 1 and 3 respectively.
Max pooling across filters is applied to get the final output $\mathbf{p}_{idc}$.

\item \textbf{N-gram Statistics\ }
To explicitly incorporate collocation information,
we use this module to output logarithmic $n$-gram counts $\mathbf{p}_{ng}$
from \emph{English Wikipedia} with $n \in [1, 5]$.
Logarithmic function avoids the optimization difficulty with extremely large numbers.
\end{itemize}

Note that the output $\mathbf{p}_{sc}$ from selective copying module
and $\mathbf{p}_{idc}$ from iterative dilated convolution module don't depend on the candidates.
We therefore get context representation $\mathbf{P}_{ctx} = [\mathbf{p}_{sc}; \mathbf{p}_{idc}]$
by concatenating $\mathbf{p}_{sc}$ and $\mathbf{p}_{idc}$.
Similarly,
we can get the representation vector $\mathbf{C}_i$ for $i$th candidate
by concatenating the output $\mathbf{p}_{ar}^i$ from attentive reader module,
$\mathbf{p}_{ng}^i$ from n-gram statistics
and $\mathbf{u}_i$ from the candidate encoder:
$\mathbf{C}_i = [\mathbf{u}_i; \mathbf{p}_{ar}^i; \mathbf{p}_{ng}^i], i \in [1, |c|]$.
\newline

\noindent
\textbf{Output Layer\ }
We use a one-timestep pointer network ~\cite{vinyals2015pointer}
to choose the correct answer from $|c|$ candidates $\{\mathbf{c}_i\}_{i=1}^{|c|}$.
Given context representation $\mathbf{P}_{ctx}$ and candidates representation $\{\mathbf{C}_i\}_{i=1}^{|c|}$,
we first refine candidates representation with a gating mechanism:
\begin{equation}
\begin{aligned}
    & \mathbf{g}_i = \sigma(\mathbf{W}_1\mathbf{P}_{ctx} + \mathbf{W}_2\mathbf{C}_i + \mathbf{b}),\ i \in [1, |c|] \\
    & \mathbf{C'}_i = \mathbf{C}_i \odot \mathbf{g}_i,\ i \in [1, |c|]
\end{aligned}
\end{equation}
$\sigma$ is the sigmoid function and $\odot$ denotes pointwise multiplication.
Then we calculate the distribution of being the correct answer over candidates with bilinear function:
\begin{equation}
    \hat{y}_i = \mathrm{softmax}_i(\mathbf{C'}_i^T \mathbf{W}_o \mathbf{P}_{ctx} + \mathbf{b}_o^T\mathbf{C'}_i),\ i \in [1, |c|]
\end{equation}

\noindent
$\{\hat{y}_i\}_{i=1}^{|c|}$ is a probability distribution and the pointer points to the candidate $\mathrm{arg\,max}_i(\hat{y}_i)$.
\newline

\noindent
\textbf{Model Learning\ }
The model is trained by minimizing the standard cross-entropy loss.

\noindent
\textbf{Discussion\ }
Different aggregation modules summarize context from different perspectives.
In order to precisely locate the correct answer,
a set of complementary aggregation modules are preferred
where one module may only focus on lexical collocation
and another module may be sensitive to the global matching.
It is worth noting that our MPNet framework can be easily extended by adding other effective aggregation modules.

In addition,
the main idea of MPNet is to some extent connected with the mixture of experts (MoE) ~\cite{masoudnia2014mixture}.
If each aggregation module can be seen as an expert,
then multiple aggregation modules become MoE.
One key difference is that aggregation modules in MPNet have heterogeneous network structures
while traditional MoE models usually consist of homogeneous experts.

\subsection{SemiMPNet: Semi-supervised Learning with Distribution Matching} \label{section:semi}

SemiMPNet is the semi-supervised variant of our proposed MPNet in Section ~\ref{section:mpnet},
with exactly the same network architecture.
Though CLOTH consists of nearly $100k$ questions,
it is generally not enough to train large neural models.
Semi-supervised learning comes to the rescue.
We propose to sample from unlabeled text to construct training examples automatically.
In order to train effectively,
we need to make the automatically generated data similar to labeled data and
ensure that candidates should have a similar distribution in original labeled data to that in the generated data.
Then,
we formulate candidates distribution matching in two datasets as two sampling problems as follows:
\newline

\noindent
\textbf{How to sample positive candidates?\ }
We assume $\emph{D}_u$ is a collection of unlabeled documents,
$\emph{D}_c$ is the collection of all candidates in the CLOTH dataset
and $V$ is the vocabulary which is composed of all the candidates occurring in CLOTH.
Each word $w_i \in V$ is associated with an unknown sampling probability $p(w_i)$.
To match the distribution of candidates between $\emph{D}_u$ and $\emph{D}_c$,
the following constraints about $p(w_i), i \in [1, |V|]$ should be satisfied:
\begin{equation} \label{equation:pos}
\begin{aligned}
    \frac{p(w_i)\#(w_i, \emph{D}_u)}{\sum_{j=1}^{|V|}p(w_j)\#(w_j, \emph{D}_u)} & =
                \frac{\#(w_i, \emph{D}_c)}{\sum_{j=1}^{|V|}\#(w_j, \emph{D}_c)}, \ i \in [1, |V|] \\
    0 \le p(w_i) \le 1, & \ i \in [1, |V|] \\
    \max\ \{p(w_i)|\ i \in &\ [1, |V|]\} = 1
\end{aligned}
\end{equation}
Function $\#(w_i, \emph{D})$ returns the frequency of $w_i$ in corpus $\emph{D}$.
The second constraint is to make sure $\{p(w_i)\}_{i=1}^{|V|}$ is a valid probability distribution
and the third constraint is to make full use of data.
There is generally no exact solution to Equation(\ref{equation:pos})
as $\#(w_i, \emph{D}_u) = 0$ and $\#(w_i, \emph{D}_c) > 0$ may hold for some $i$.
Instead,
we use an approximate solution:
\begin{equation}
p(w_i) = \min(1, \frac{\#(w_i, \emph{D}_c)}{\#(w_i, \emph{D}_u)} \times
        \frac{\gamma \sum_{j=1}^{|V|}\#(w_j, \emph{D}_u)}{\sum_{j=1}^{|V|}\#(w_j, \emph{D}_c)})
\end{equation}
The coefficient $\gamma$ can be interpreted as the average probability of sampling a word.
We set $\gamma = 0.5$ based on validation data.
With this strategy,
we sample the positive candidates and use the corresponding passages as their contexts.

\noindent
\textbf{How to sample negative candidates?\ }
Given a positive candidate $w_p$,
the probability of $w_i$ being sampled as a negative candidate $p(w_i | w_p)$ can be calculated as follows:
\begin{equation}
p(w_i | w_p) = \frac{\lambda}{|V|} + (1-\lambda) \frac{\#(w_i, w_p)}{\sum_{j=1}^{|V|}\#(w_j, w_p)}
\end{equation}
$\#(w_i, w_p)$ is the co-occurrence counts of $w_i$ and $w_p$ as candidates in labeled dataset $\emph{D}_c$.
Intuitively,
the co-occurrence probability of $w_i$ and $w_p$ should match between constructed data and labeled data.
$\lambda$ is the probability of randomly selecting a word from the entire vocabulary,
similar to the exploration mechanism in reinforcement learning.
It makes our model more robust to overfitting
and we set $\lambda = 0.1$ throughout the experiments.

In the case that candidates are multi-word phrases,
our method is also applicable
by simply expanding the vocabulary $V$ to include phrases in $\emph{D}_c$.

\section{Experiments}

\subsection{Experimental Setup}
\noindent
\textbf{Dataset and Evaluation Metrics\ }
We use the CLOTH~\cite{xie2017large} dataset for training and evaluation.
RACE~\cite{lai2017race} dataset and \emph{English Wikipedia}~\footnote{\url{https://dumps.wikimedia.org/enwiki/}}
serve as background text corpora for semi-supervised learning.
RACE dataset consists of nearly $28k$ reading comprehension passages from high-school examinations.
We delete passages
that have a Jaccard similarity over $0.85$ with passages in the CLOTH dataset.
Furthermore,
background text corpora also include training passages from the CLOTH dataset
by filling the correct answer back into the corresponding blank.

Accuracy is used as the evaluation metric.
To make a fair comparison with ~\newcite{xie2017large},
we also report performance on CLOTH-M(middle school questions) and CLOTH-H(high school questions).

\noindent
\textbf{Hyperparameters\ }
Our model is implemented with Tensorflow ~\cite{abadi2016tensorflow}.
Hyperparameters are optimized with random search based on validation data.
All our models are run on a single GPU(Tesla P40).
NLTK ~\cite{bird2004nltk} is used for tokenization.
Word embeddings are initialized with 300-dimensional GloVe ~\cite{pennington2014glove} vectors.
Only vectors of top $1000$ frequent words are fine-tuned during training.
Our network is trained with Adam algorithm ~\cite{kingma2014adam}.
The initial learning rate is set to $10^{-3}$.
We decrease learning rate to $10^{-4}$ after $15k$ iterations
and further decrease it to $10^{-5}$ after $50k$ iterations.
Both forward and backward GRU have $128$ hidden units.
For input,
we use a context window of $80$ words.
For 1D dilated convolution,
we use $2$ blocks,
the number of filters is $128$
and the convolution width is $3$ for all layers.
Batch normalization and ReLU are applied on top of convolution.
Gradients are clipped to have a maximum L2 norm of $5$.
Dropout with probability $0.5$ is applied to the output of BiGRU.

\begin{table*}[ht]
\centering
\scalebox{0.95}{\begin{tabular}{c|c|ccc}
 \hline
Model                   & + constructed data? & CLOTH & CLOTH-M & CLOTH-H \\ \hline
\emph{Random}     & No              & 25.0\%     & 25.0\%   & 25.0\%  \\
\emph{LSTM}~\cite{xie2017large}     & No              & 48.4\%     & 51.8\%   & 47.1\%  \\
\emph{Stanford Attention Reader}~\cite{chen2016thorough}    & No      &   48.7\%     & 52.9\%   & 47.1\%  \\
\emph{MPNet - ngram}                & No                   & \textbf{50.1\%}  & \textbf{53.2\%} & \textbf{49.0\%} \\ \hline \hline
\emph{Language Model} ~\cite{xie2017large}   & Yes    & 54.8\%  & 64.6\% & 50.6\%   \\
\emph{Representativeness} ~\cite{xie2017large} & Yes & 56.5\% & 66.5\% & 52.6\% \\
\emph{LSTM + Representativeness} ~\cite{xie2017large} & Yes & 58.3\%  & 67.3\%  & 54.9\%  \\
\emph{SemiMPNet - ngram} & Yes      & \textbf{60.9\%} & \textbf{67.6\%}  & \textbf{58.3\%} \\ \hline \hline
\emph{Human} & -- & 86.0\%  &  89.7\%  & 84.5\%   \\ \hline
\end{tabular}}
\caption{Experimental results without using external data.
We exclude \emph{n-gram} as \emph{n-gram} is calculated based on external corpus \emph{Wikipedia}.
SemiMPNet uses passages from CLOTH for semi-supervised data augmentation.
Human performance is from \protect\newcite{xie2017large}.}
\label{table:main_results}
\end{table*}

\subsection{Baselines}

\noindent
\textbf{LSTM\ }
is a baseline model by ~\newcite{xie2017large}.
First,
a BiLSTM layer is applied to context word embeddings.
Then it uses the outputs near the blank to calculate the probability
of being the correct answer for each candidate.

\noindent
\textbf{Stanford Attention Reader\ }
is an attention-based neural model for reading comprehension
presented by ~\newcite{chen2016thorough}.
Experimental results are from ~\newcite{xie2017large}.

\noindent
\textbf{Language Model\ }
To overcome the difficulty of insufficient labeled data.
~\newcite{xie2017large} propose to train a neural language model
on passages from the CLOTH dataset.
The candidate that results in the highest probability is chosen as the predicted answer.
It's fair to say
\emph{Language Model} is a simple data augmentation approach
that treats every word as a training example with equal weight.

\noindent
\textbf{Representativeness\ }
is another semi-supervised data augmentation approach by ~\newcite{xie2017large}.
It assigns different weights to different constructed examples based on \emph{Representativeness score}.
\emph{Representativeness} can be interpreted as
the probability of a given word being selected as a blank by human.
For more technical details,
please refer to ~\newcite{xie2017large}.

\noindent
\textbf{One-billion-word-LM\ }
is a state-of-the-art neural language model~\cite{jozefowicz2016exploring}
trained on one-billion-word benchmark~\cite{chelba2013one}.
It has more than $1$ billion parameters
and is publicly available\footnote{\url{https://github.com/tensorflow/models/tree/master/research/lm_1b}}.

\subsection{Main Results}
We evaluate our model's performance in two experimental settings:
use external data or not.
For the setting without external data,
we only use passages from CLOTH for training and semi-supervised data augmentation.
Though GloVe vectors are trained on external text corpora,
it has become a standard practice for NLP to use pretrained embeddings.
Therefore,
GloVe vectors are used in both settings
and so does the work by ~\newcite{xie2017large}.
\newline

\noindent
\textbf{Results w/o External Data\ }
Results are shown in Table ~\ref{table:main_results}.
When trained only on labeled data,
both LSTM by ~\newcite{xie2017large} and our proposed MPNet perform poorly,
though MPNet slightly outperforms LSTM by $1.7\%$ in overall accuracy.
The accuracy of middle school questions (CLOTH-M) is
consistently higher than high school questions (CLOTH-H) across all of our experiments,
since middle school questions are relatively easier.

\begin{table}[ht]
\centering
\begin{tabular}{cc|cc|cc}
 \hline
$w$     & $p(w)$ & $w$    & $p(w)$ & $w$         & $p(w)$ \\ \hline
I     & $0.04$ & festivals & $0.75$ & California     & $0.13$ \\
the   & $0.03$ & birthday  & $1.0$  & thank you      & $0.26$ \\
Frank & $0.09$ & 8         & $0.09$ & congratulation & $1.0$    \\ \hline
\end{tabular}
\caption{Sample probability for some words.
$p(w)$ is the probability of sampling $w$ to construct a training example.
Stop words, named entities usually have low probability.
See Section ~\ref{section:semi} for details.}
\label{table:sample_proba}
\end{table}

Table ~\ref{table:main_results} clearly shows that
constructed data can significantly boost both models' performance.
~\newcite{xie2017large} explore several different ways for data augmentation:
\emph{Language Model} treats every word equally,
while \emph{Representativeness} method assigns different weights to different words
by training an representativeness prediction network.
This mechanism improves the accuracy from $48.4\%$ to $58.3\%$.
Further, our proposed method adopts a new sampling method and requires sampling words with distribution constraints,
which makes training more efficient.
As shown in Table ~\ref{table:sample_proba},
stop words (e.g., \emph{``I''} and \emph{``the''}),
named entities (e.g., \emph{``Frank''} and \emph{``California''})
and common phrases (e.g., \emph{``thank you''}) have low probability of being sampled.
Content words such as \emph{``festivals''} and \emph{``birthday''} are more likely to be sampled.
One limitation of our sampling method is its inability to handle synonyms.
Since synonyms tend to co-occur as candidates in the labeled dataset,
this problem is not as severe as it looks like to be.
\emph{``SemiMPNet - ngram''} beats all baseline methods
and achieves the highest accuracy $60.9\%$.
Human performance is $86.0\%$ which is much higher than \emph{``SemiMPNet - ngram''}.
The effectiveness of constructed data indicates that
the lack of labeled data has become a bottleneck.
\newline

\begin{table}[ht]
\centering
\scalebox{0.92}{\begin{tabular}{c|ccc}
 \hline
Model                & CLOTH & CLOTH-M & CLOTH-H \\ \hline
\emph{One-billion-word-LM} & 70.7\%    & 74.5\%   & 69.3\%   \\
\emph{MPNet}         & 65.3\%  & 70.0\%    & 63.6\%       \\
\emph{SemiMPNet}         & 70.4\%     & 75.5\%  & 68.5\%  \\
\emph{SemiMPNet + One-billion-word-LM} & \textbf{74.9\%}  & \textbf{79.0\%}  & \textbf{73.3\%} \\ \hline \hline
\emph{Human} & 86.0\%  &  89.7\%  & 84.5\%   \\ \hline
\end{tabular}}
\caption[filler]{Experimental results with external data.
SemiMPNet use passages from the RACE dataset for semi-supervised data augmentation.}
\label{table:semi_supervised}
\end{table}

\noindent
\textbf{Results with External Data\ }
As shown in Table ~\ref{table:semi_supervised},
incorporation of the RACE dataset for semi-supervised learning
improves accuracy from $65.3\%$ to $70.4\%$.
However,
MPNet and SemiMPNet still underperform
a pretrained state-of-the-art neural language model \emph{One-billion-word-LM}~\cite{jozefowicz2016exploring}.
It is trained on a large corpus with nearly $1$ billion words
and achieves an accuracy of $70.7\%$.
In contrast,
SemiMPNet is trained on only $\sim10$ million words
and has a $0.3\%$ gap in accuracy,
which is pretty impressive given that the sizes of two corpora differ by two orders of magnitude.
Once again it shows the power of transferring knowledge from unlabeled text corpora.

As a further discussion,
we'd like to point out that although language model can achieve good results,
it is not the most efficient way.
Actually,
experimental results in Table ~\ref{table:main_results} show that
language model underperforms SemiMPNet given the same amount of text.
A fair comparison would be training SemiMPNet on the one-billion-word benchmark.
Considering the size of the one-billion-word corpus,
applying our semi-supervised method directly on the one-billion-word corpus
would require a sizable amount of computing power.
Here we design an approximate method \emph{``SemiMPNet + One-billion-word-LM''}
and combine MPNet and \emph{One-billion-word-LM} by linear interpolation of their output probabilities:
\begin{equation}
p = \beta p_{mp} + (1 - \beta) p_{lm}
\end{equation}
$p_{mp}$ is the output probability by SmiMPNet
and $p_{lm}$ is the normalized probability by
\emph{One-billion-word-LM}\footnote{The normalization makes sure the probabilities for all candidates sum to 1.}.
Setting $\beta = 0.5$ yields empirically good results.
Hyper-parameter search shows the results are quite robust to a wide range of $\beta$ values.
We can see that our model \emph{``SemiMPNet + One-billion-word-LM''} achieves a new state-of-the-art performance of $74.9\%$,
which improves \emph{One-billion-word-LM} by $4.2\%$.
This also shows the complementarity of SemiMPNet and \emph{One-billion-word-LM}.
Two models can learn different aspects of the contexts.

\subsection{Ablation Study}
Our proposed MPNet consists of four aggregation modules.
To examine the effect of each module,
we conduct an ablation study.
The results are shown in Table ~\ref{table:ablation}.

\begin{table}[ht]
\centering
\begin{tabular}{l|c}
 \hline
\multicolumn{1}{c|}{Model} & CLOTH \\ \hline
\multicolumn{1}{c|}{\emph{SemiMPNet}} & \textbf{70.4\%} \\ \hline
\emph{w/o selective copying}        & 69.4\% (-1.0)     \\
\emph{w/o attentive reader}         & 67.6\% (-2.8)     \\
\emph{w/o dilated convolution}      & 69.6\% (-0.8)     \\
\emph{w/o n-gram statistics}        & 63.0\% (-7.4)    \\ \hline
\end{tabular}
\caption{Results for SemiMPNet ablation study.
RACE dataset is used for semi-supervised data augmentation.}
\label{table:ablation}
\end{table}

\emph{N-gram statistics} turn out to be the single most influential factor.
Overall performance decreases by $7.4\%$ without \emph{n-gram statistics}.
On one hand,
this result further highlights the importance of distilling knowledge from large text corpora.
On the other hand,
it proves that our background corpus is not large enough
for neural models to learn reliable lexical collocation information.

\emph{Attentive reader} also has a significant impact on overall performance.
Attention mechanism is able to locate useful information
regardless of its positional distance from the blank.
In contrast,
RNNs need to preserve such information over a long distance which is nontrivial.

Besides,
the results in Table ~\ref{table:ablation} support an important intuition in this paper:
different modules capture context information from different perspectives,
and removing any one of them would result in decreased performance.

\begin{figure}[ht]
\begin{center}
 \includegraphics[width=0.5\linewidth]{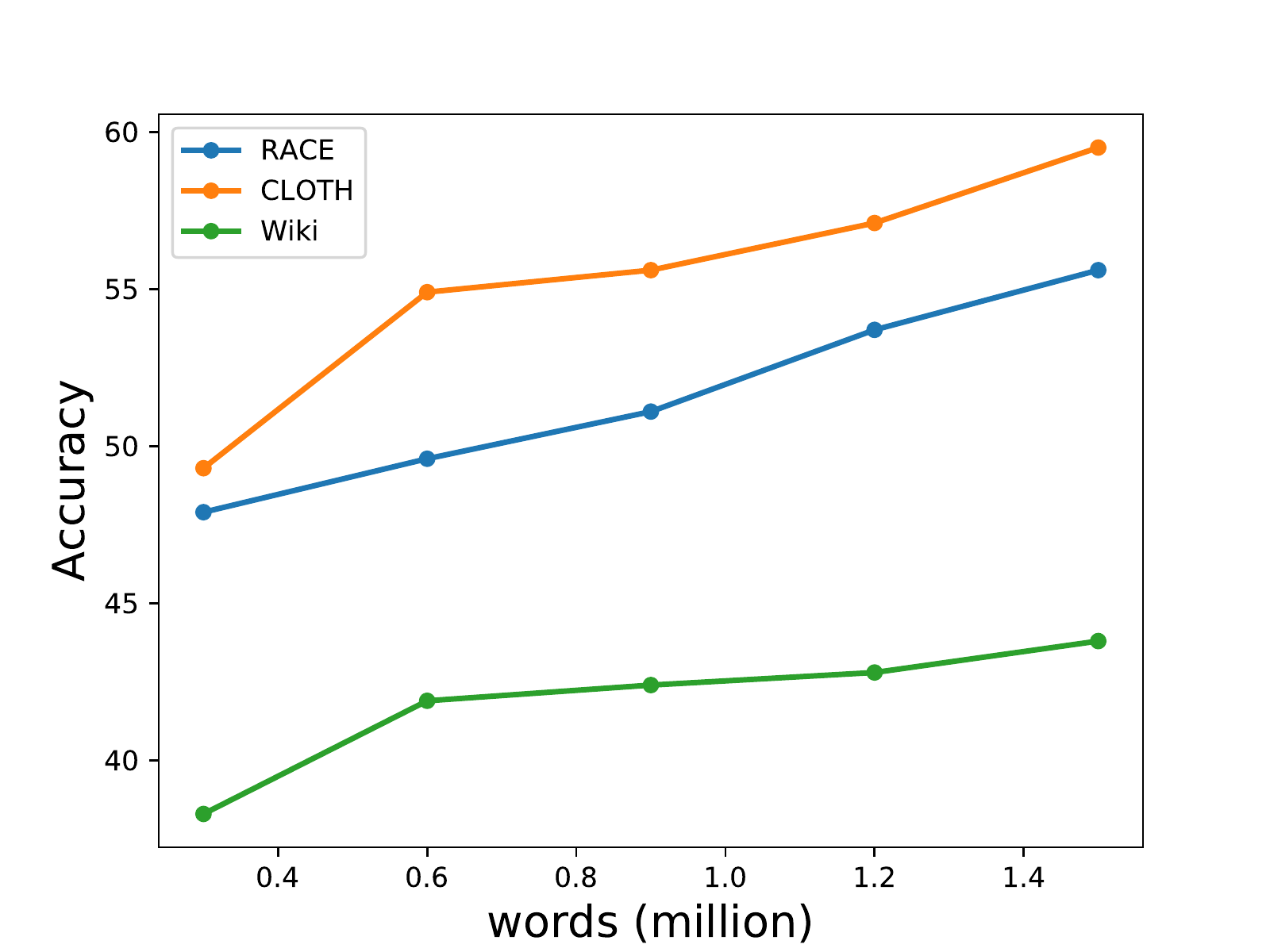}
 \caption{Examining effects of different text corpora.
 The x-axis is the number of words in background corpus,
 and the y-axis is the accuracy on test data.
 To avoid the influence of external data,
 we report model performance with \emph{``SemiMPNet - ngram''}.}
 \label{fig:corpus}
\end{center}
\end{figure}

\subsection{Examining Effects of Background Corpus}

For our semi-supervised learning model SemiMPNet,
background corpus is used to construct training examples.
The choice of background corpus can make a big difference.
In this section,
we conduct an experiment to examine such effects.
Three different corpora are used:
passages from the training set of CLOTH,
RACE and English Wikipedia.

Results are shown in Figure ~\ref{fig:corpus}.
Unsurprisingly,
more data lead to better performance.
Moreover,
given the same amount of text,
CLOTH consistently beats RACE
and RACE consistently beats Wikipedia.
As we know,
CLOTH and RACE consists of passages designed for high school students,
while Wikipedia entries are for the general public
and therefore have a different word distribution.
Thus, how to make use of huge unlabeled data to help training is a key for performance improvement,
since training corpora of higher quality are generally smaller in scale.

\section{Related Work}

\textbf{Reading Comprension} or machine reading is drawing more and more interests among NLP research communities.
The CNN/Daily Mail ~\cite{hermann2015teaching} and CBT ~\cite{hill2015goldilocks}
are two automatically generated cloze-style datasets.
Though they can be large in scale,
the quality of automatically generated questions is generally lower than manually labeled ones.
Instead,
SQuAD~\cite{rajpurkar2016squad} adopts a crowd-sourcing approach to ensure its quality.
In SQuAD,
each passage accompanies one or more questions
and the answer is a text span of the given passage
for the convenience of automatic evaluation.
Rapid progress has been made with neural network based models ~\cite{wang2016multi}.
The performances of state-of-the-art models on SQuAD
such as QANet ~\cite{yu2018qanet} and ELMo ~\cite{peters2018deep} are already very close to human.
There are also some datasets focusing on answering questions
from real-world scenarios.
MS MARCO ~\cite{nguyen2016ms} and DuReader ~\cite{he2017dureader} are two typical examples.
Such datasets are usually harder as they require the ability of
both comprehension and language generation.
BLEU and ROUGE are often used as evaluation metrics.
One potential problem is
that answers with high BLEU scores may have very different semantic meanings.
\newline

\noindent
\textbf{Cloze Test} is a particular form of reading comprehension task
and has been widely adopted as a method for assessing students' language proficiency.
~\newcite{zweig2011microsoft} presented a challenging dataset for sentence completion
but its scale is too small with only $1040$ questions.
CNN/Daily Mail ~\cite{hermann2015teaching}, CBT ~\cite{hill2015goldilocks}, LAMBADA~\cite{paperno2016lambada}
and CLOTH ~\cite{xie2017large} are all large-scale cloze-test datasets,
with the difference that each question in CLOTH has four candidate options.
Recently proposed Story Cloze ~\cite{mostafazadeh2017lsdsem} is a cloze-test dataset
that goes beyond words and phrases,
and requires choosing a sentence as the appropriate story ending.
\newline

\noindent
\textbf{Semi-supervised Methods} for reading comprehension are widely studied
due to the fact that labeled data is scarce.
One major approach is pretraining a model for text representation and reusing the weights during supervised learning.
Autoencoders~\cite{hewlett2017accurate}, machine translation~\cite{mccann2017learned}
and language model ~\cite{peters2018deep} can be used for representation learning.
Another approach aims to directly construct training examples from unlabeled text corpora.
Weighted loss function~\cite{xie2017large} and reinforcement learning~\cite{yang2017semi} can be used
to alleviate the discrepancy between human-labeled data and automatically-constructed data.

\section{Conclusion}
In this paper,
we propose a multi-perspective network MPNet for cloze-style reading comprehension.
MPNet consists of several parallel context aggregation modules.
Each module summarizes the variable-length context and candidates into a fixed-length vector from a unique perspective.
We explore four effective implementations of aggregation modules in experiments.
The architecture of MPNet is very flexible
and can be easily extended by adding more task-specific modules.

To overcome the difficulty of limited labeled data,
we turn to semi-supervised learning by automatically constructing training examples from unlabeled text corpora.
Experiments on the CLOTH dataset show
that our semi-supervised MPNet achieves new state-of-the-art performance.
In our future work,
we'd like to come up with more effective methods to tackle this challenge.

\section*{Acknowledgements}
We would like to thank three anonymous reviewers for their insightful comments,
and COLING 2018 organizers for their efforts.

\bibliographystyle{acl}
\bibliography{coling2018}

\begin{thebibliography}{}

\bibitem[\protect\citename{Abadi \bgroup et al.\egroup
  }2016]{abadi2016tensorflow}
Mart{\'\i}n Abadi, Paul Barham, Jianmin Chen, Zhifeng Chen, Andy Davis, Jeffrey
  Dean, Matthieu Devin, Sanjay Ghemawat, Geoffrey Irving, Michael Isard, et~al.
\newblock 2016.
\newblock {TensorFlow}: A system for large-scale machine learning.
\newblock In {\em OSDI}, volume~16, pages 265--283.

\bibitem[\protect\citename{Bird and Loper}2004]{bird2004nltk}
Steven Bird and Edward Loper.
\newblock 2004.
\newblock {NLTK: the natural language toolkit}.
\newblock In {\em Proceedings of the ACL 2004 on Interactive poster and
  demonstration sessions}, page~31. Association for Computational Linguistics.

\bibitem[\protect\citename{Chelba \bgroup et al.\egroup }2013]{chelba2013one}
Ciprian Chelba, Tomas Mikolov, Mike Schuster, Qi~Ge, Thorsten Brants, Phillipp
  Koehn, and Tony Robinson.
\newblock 2013.
\newblock One billion word benchmark for measuring progress in statistical
  language modeling.
\newblock {\em arXiv preprint arXiv:1312.3005}.

\bibitem[\protect\citename{Chen \bgroup et al.\egroup }2016]{chen2016thorough}
Danqi Chen, Jason Bolton, and Christopher~D Manning.
\newblock 2016.
\newblock {A Thorough Examination of the CNN/Daily Mail Reading Comprehension
  Task}.
\newblock In {\em Proceedings of the 54th Annual Meeting of the Association for
  Computational Linguistics (Volume 1: Long Papers)}, volume~1, pages
  2358--2367.

\bibitem[\protect\citename{He \bgroup et al.\egroup }2017]{he2017dureader}
Wei He, Kai Liu, Yajuan Lyu, Shiqi Zhao, Xinyan Xiao, Yuan Liu, Yizhong Wang,
  Hua Wu, Qiaoqiao She, Xuan Liu, et~al.
\newblock 2017.
\newblock {DuReader: a Chinese Machine Reading Comprehension Dataset from
  Real-world Applications}.
\newblock {\em arXiv preprint arXiv:1711.05073}.

\bibitem[\protect\citename{Hermann \bgroup et al.\egroup
  }2015]{hermann2015teaching}
Karl~Moritz Hermann, Tomas Kocisky, Edward Grefenstette, Lasse Espeholt, Will
  Kay, Mustafa Suleyman, and Phil Blunsom.
\newblock 2015.
\newblock Teaching machines to read and comprehend.
\newblock In {\em Advances in Neural Information Processing Systems}, pages
  1693--1701.

\bibitem[\protect\citename{Hewlett \bgroup et al.\egroup
  }2017]{hewlett2017accurate}
Daniel Hewlett, Llion Jones, Alexandre Lacoste, et~al.
\newblock 2017.
\newblock Accurate supervised and semi-supervised machine reading for long
  documents.
\newblock In {\em Proceedings of the 2017 Conference on Empirical Methods in
  Natural Language Processing}, pages 2001--2010.

\bibitem[\protect\citename{Hill \bgroup et al.\egroup
  }2015]{hill2015goldilocks}
Felix Hill, Antoine Bordes, Sumit Chopra, and Jason Weston.
\newblock 2015.
\newblock The goldilocks principle: Reading children's books with explicit
  memory representations.
\newblock {\em arXiv preprint arXiv:1511.02301}.

\bibitem[\protect\citename{Jozefowicz \bgroup et al.\egroup
  }2016]{jozefowicz2016exploring}
Rafal Jozefowicz, Oriol Vinyals, Mike Schuster, Noam Shazeer, and Yonghui Wu.
\newblock 2016.
\newblock Exploring the limits of language modeling.
\newblock {\em arXiv preprint arXiv:1602.02410}.

\bibitem[\protect\citename{Kim}2014]{kim2014convolutional}
Yoon Kim.
\newblock 2014.
\newblock Convolutional neural networks for sentence classification.
\newblock {\em arXiv preprint arXiv:1408.5882}.

\bibitem[\protect\citename{Kingma and Ba}2014]{kingma2014adam}
Diederik Kingma and Jimmy Ba.
\newblock 2014.
\newblock Adam: A method for stochastic optimization.
\newblock {\em arXiv preprint arXiv:1412.6980}.

\bibitem[\protect\citename{Lai \bgroup et al.\egroup }2017]{lai2017race}
Guokun Lai, Qizhe Xie, Hanxiao Liu, Yiming Yang, and Eduard Hovy.
\newblock 2017.
\newblock {RACE}: Large-scale reading comprehension dataset from examinations.
\newblock {\em arXiv preprint arXiv:1704.04683}.

\bibitem[\protect\citename{Masoudnia and
  Ebrahimpour}2014]{masoudnia2014mixture}
Saeed Masoudnia and Reza Ebrahimpour.
\newblock 2014.
\newblock Mixture of experts: a literature survey.
\newblock {\em Artificial Intelligence Review}, pages 1--19.

\bibitem[\protect\citename{McCann \bgroup et al.\egroup
  }2017]{mccann2017learned}
Bryan McCann, James Bradbury, Caiming Xiong, and Richard Socher.
\newblock 2017.
\newblock Learned in translation: Contextualized word vectors.
\newblock {\em arXiv preprint arXiv:1708.00107}.

\bibitem[\protect\citename{Mostafazadeh \bgroup et al.\egroup
  }2017]{mostafazadeh2017lsdsem}
Nasrin Mostafazadeh, Michael Roth, Annie Louis, Nathanael Chambers, and James~F
  Allen.
\newblock 2017.
\newblock {LSDSem 2017 Shared Task: The Story Cloze Test}.
\newblock {\em LSDSem 2017}, page~46.

\bibitem[\protect\citename{Nguyen \bgroup et al.\egroup }2016]{nguyen2016ms}
Tri Nguyen, Mir Rosenberg, Xia Song, Jianfeng Gao, Saurabh Tiwary, Rangan
  Majumder, and Li~Deng.
\newblock 2016.
\newblock {MS MARCO: A human generated machine reading comprehension dataset}.
\newblock {\em arXiv preprint arXiv:1611.09268}.

\bibitem[\protect\citename{Paperno \bgroup et al.\egroup
  }2016]{paperno2016lambada}
Denis Paperno, Germ{\'a}n Kruszewski, Angeliki Lazaridou, Quan~Ngoc Pham,
  Raffaella Bernardi, Sandro Pezzelle, Marco Baroni, Gemma Boleda, and Raquel
  Fern{\'a}ndez.
\newblock 2016.
\newblock {The LAMBADA dataset: Word prediction requiring a broad discourse
  context}.
\newblock {\em arXiv preprint arXiv:1606.06031}.

\bibitem[\protect\citename{Pennington \bgroup et al.\egroup
  }2014]{pennington2014glove}
Jeffrey Pennington, Richard Socher, and Christopher Manning.
\newblock 2014.
\newblock {GloVe: Global vectors for word representation}.
\newblock In {\em Proceedings of the 2014 conference on empirical methods in
  natural language processing (EMNLP)}, pages 1532--1543.

\bibitem[\protect\citename{Peters \bgroup et al.\egroup }2018]{peters2018deep}
Matthew~E Peters, Mark Neumann, Mohit Iyyer, Matt Gardner, Christopher Clark,
  Kenton Lee, and Luke Zettlemoyer.
\newblock 2018.
\newblock Deep contextualized word representations.
\newblock {\em NAACL}.

\bibitem[\protect\citename{Rajpurkar \bgroup et al.\egroup
  }2016]{rajpurkar2016squad}
Pranav Rajpurkar, Jian Zhang, Konstantin Lopyrev, and Percy Liang.
\newblock 2016.
\newblock Squad: 100,000+ questions for machine comprehension of text.
\newblock {\em arXiv preprint arXiv:1606.05250}.

\bibitem[\protect\citename{Strubell \bgroup et al.\egroup
  }2017]{strubell2017fast}
Emma Strubell, Patrick Verga, David Belanger, and Andrew McCallum.
\newblock 2017.
\newblock Fast and accurate entity recognition with iterated dilated
  convolutions.
\newblock In {\em Proceedings of the 2017 Conference on Empirical Methods in
  Natural Language Processing}, pages 2660--2670.

\bibitem[\protect\citename{Vinyals \bgroup et al.\egroup
  }2015]{vinyals2015pointer}
Oriol Vinyals, Meire Fortunato, and Navdeep Jaitly.
\newblock 2015.
\newblock Pointer networks.
\newblock In {\em Advances in Neural Information Processing Systems}, pages
  2692--2700.

\bibitem[\protect\citename{Wang \bgroup et al.\egroup }2016]{wang2016multi}
Zhiguo Wang, Haitao Mi, Wael Hamza, and Radu Florian.
\newblock 2016.
\newblock Multi-perspective context matching for machine comprehension.
\newblock {\em arXiv preprint arXiv:1612.04211}.

\bibitem[\protect\citename{Xie \bgroup et al.\egroup }2017]{xie2017large}
Qizhe Xie, Guokun Lai, Zihang Dai, and Eduard Hovy.
\newblock 2017.
\newblock Large-scale cloze test dataset designed by teachers.
\newblock {\em arXiv preprint arXiv:1711.03225}.

\bibitem[\protect\citename{Yang \bgroup et al.\egroup }2017]{yang2017semi}
Zhilin Yang, Junjie Hu, Ruslan Salakhutdinov, and William~W Cohen.
\newblock 2017.
\newblock {Semi-Supervised QA with Generative Domain-Adaptive Nets}.
\newblock {\em arXiv preprint arXiv:1702.02206}.

\bibitem[\protect\citename{Yu and Koltun}2015]{yu2015multi}
Fisher Yu and Vladlen Koltun.
\newblock 2015.
\newblock Multi-scale context aggregation by dilated convolutions.
\newblock {\em arXiv preprint arXiv:1511.07122}.

\bibitem[\protect\citename{Yu \bgroup et al.\egroup }2018]{yu2018qanet}
Adams~Wei Yu, David Dohan, Minh-Thang Luong, Rui Zhao, Kai Chen, Mohammad
  Norouzi, and Quoc~V Le.
\newblock 2018.
\newblock Qanet: Combining local convolution with global self-attention for
  reading comprehension.
\newblock {\em ICLR}.

\bibitem[\protect\citename{Zweig and Burges}2011]{zweig2011microsoft}
Geoffrey Zweig and Christopher~JC Burges.
\newblock 2011.
\newblock {The Microsoft Research sentence completion challenge}.
\newblock Technical report, Technical Report MSR-TR-2011-129, Microsoft.

\end{thebibliography}

\end{document}